\documentclass[english]{article}
\usepackage[preprint,nonatbib]{nips_2018_wider_nonotice}

\usepackage[T1]{fontenc}
\usepackage[latin9]{inputenc}
\usepackage{color,colortbl}
\usepackage{babel}
\usepackage{verbatim}
\usepackage{url}
\usepackage{amsmath}
\usepackage{amssymb}
\usepackage{graphicx}
\usepackage{setspace}
\usepackage{hyperref}
\usepackage{cancel}
\hypersetup{unicode=true,breaklinks=false}
\usepackage[hyperpageref]{backref}
\usepackage{appendix}
\usepackage{xspace}
\usepackage{epigraph}
\usepackage[thinc]{esdiff}
\usepackage{graphicx}
\usepackage{subfigure}
\usepackage[capitalize]{cleveref}
\graphicspath{{figs/}}

\usepackage[labelfont=bf]{caption}
\makeatletter
\newcommand*{\rom}[1]{\expandafter\@slowromancap\romannumeral #1@}
\makeatother

\makeatletter
\newcommand{\be}{\begin{equation}}
\newcommand{\ee}{\end{equation}}

\allowdisplaybreaks \numberwithin{equation}{section}
\makeatother

\hypersetup{
    colorlinks=true,
    breaklinks=true,
    urlcolor=blue,
    linkcolor=blue,
    bookmarksopen=false,
    filecolor=black,
    citecolor=blue,
}

\title{Exploring Scaling Laws for Local SGD in Large Language Model Training}

\author{
    Qiaozhi He\thanks{Equal contribution. correspondence to qiaozhihe2022@outlook.com}
    \And
    Xiaomin Zhuang$^{\ast}$
    \And
    Zhihua Wu
    \AND
\\
}

\begin{document}
\maketitle

\begin{abstract}
This paper investigates scaling laws for local SGD \cite{stich2019localsgdconvergesfast} in LLM training, a distributed optimization algorithm that facilitates training on loosely connected devices. Through extensive experiments, we show that local SGD achieves competitive results compared to conventional methods, given equivalent model parameters, datasets, and computational resources.
Furthermore, we explore the application of local SGD in various practical scenarios, including multi-cluster setups and edge computing environments. Our findings elucidate the necessary conditions for effective multi-cluster LLM training and examine the potential and limitations of leveraging edge computing resources in the LLM training process. This demonstrates its viability as an alternative to single large-cluster training.
\end{abstract}

\newpage

\section{Introduction}
\begin{figure} 
\noindent \centering{}
\includegraphics[width=0.48\textwidth]{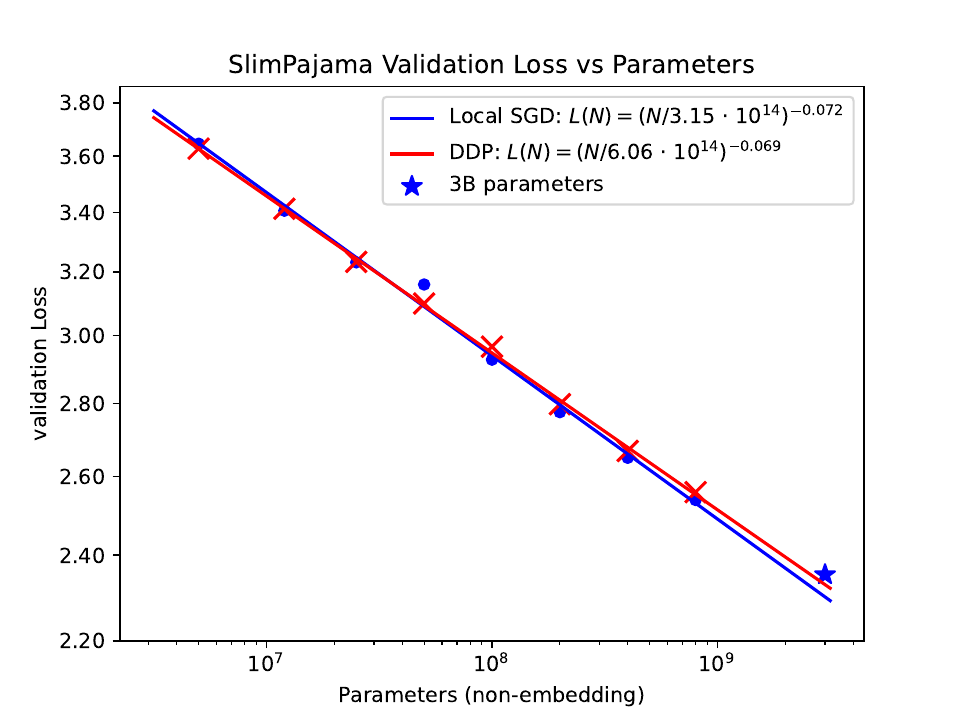}
\includegraphics[width=0.48\textwidth]{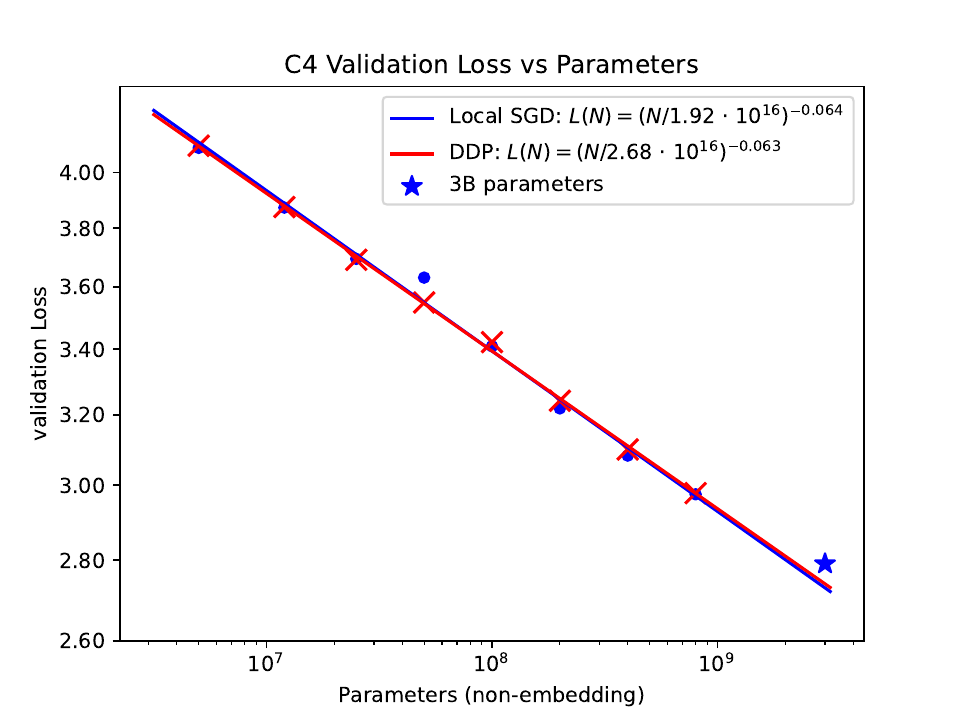}
 \caption{Scaling laws for local SGD in Large Language Models. The validation results on the SlimPajama datasets are presented on the left, whereas the right side displays the out-of-distribution results from the validation on the C4 datasets.}
 \label{fig:main_loss}
\end{figure}

Large language models (LLMs) \cite{dubey2024llama3herdmodels,gemmateam2024gemma2improvingopen,yang2024qwen2technicalreport} have demonstrated exceptional performance through training on extensive datasets by utilizing large-scale clusters. \cite{kaplan2020scalinglawsneurallanguage,hoffmann2022trainingcomputeoptimallargelanguage} has elucidated the scaling law of LLMs, which articulates the power-law relationships between model performance and variables such as computational power, parameter count, and data size. This scaling law serves as a valuable framework for forecasting model performance and planning LLM training. 

According to our limited knowledge, no algorithmic bottlenecks in these scaling laws have been identified. However, empirical evidence from open-source models and public data indicates a deceleration in the exponential growth of model size and data throughput. One of the main factors contributing to this slowdown is the escalating demand for computing resources. While \cite{narayanan2021efficientlargescalelanguagemodel} proposes a Megatron-LM training framework to guide efficient LLM distributed training implementations, these typically require high-bandwidth, non-blocking networks. The infrastructure challenges associated with scaling include: Small clusters require just a layer or two of switches, whereas large GPU clusters need more layers of switches, increasing construction costs. Many existing computational clusters have limited scalability due to constraints in air conditioning, network connectivity, power supply, and structural load capacity. This impedes the short-term scalability of existing GPU clusters. Establishing ultra-large computational clusters at a single location poses significant electrical power challenges. These three factors constrain the further expansion of the LLM scale.

Exploring physically distributed multi-cluster configurations to mitigate the dependence on single large-cluster training for LLMs presents a promising avenue. Local SGD \cite{stich2019localsgdconvergesfast} proposed a way to reduce the frequency of communication to overcome the communication bottleneck. This method has demonstrated promising results in federated learning, and \cite{douillard2023dilocodistributedlowcommunicationtraining} has shown its feasibility in LLM training. However, there is a lack of evidence that this approach is scalable. The viability of this approach hinges on addressing two key questions: Can LLMs maintain comparable scaling capabilities to traditional methods in this distributed scenario? What is the scaling potential of this method given current mainstream hardware configurations and network bandwidth limitations? By addressing these questions, we evaluated the feasibility of distributed multi-cluster approaches as a potential solution to the scaling challenges faced in LLM training. This exploration offers new pathways for advancing LLM development beyond the constraints of single large-scale clusters.

In this paper, we propose scaling laws for local SGD in large language model training. We conduct extensive experiments to investigate the scaling law of this new optimization algorithm and compare it with traditional model training methods. Finally, we consider multiple practical scenarios, such as multiple clusters, edge computing, etc., and give the scaling law of the local SGD algorithm in the LLM training process. This provides a basis for the possible training of LLM on multiple clusters in the future and discusses the possibilities and limitations of edge computing in the LLM training process.

In summary, the contributions of our work are as follows: 
\begin{itemize}
\item We demonstrate that LLM training based on local SGD exhibits scaling law. This law provides a framework for understanding and predicting performance in distributed training environments.
\item Our extensive experiments reveal that local SGD achieves promising results compared to traditional data parallelism schemes, given equivalent model parameter scales, datasets, and computational resources.
\item We elucidate the necessary conditions for effective multi-cluster training of LLMs, offering a potential pathway for further scaling of these models.
\item Our analysis explores the possibilities and limitations of edge computing in LLM training, providing valuable insights for leveraging distributed computational resources.
\end{itemize}

\begin{figure} 
\noindent \centering{} 
\includegraphics[width=6in]{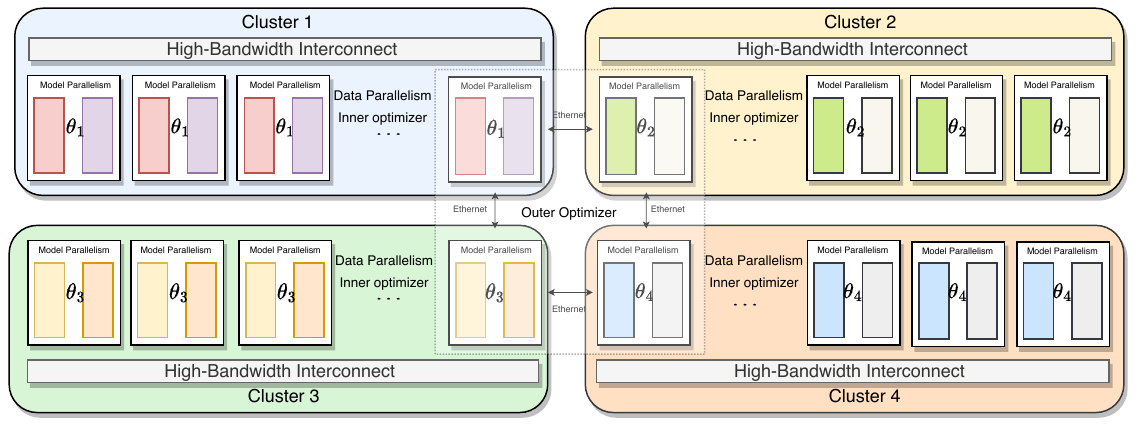} 
 \caption{A depiction of the local SGD training procedure on cross-regional clusters.}
 \label{fig:local_llm}
\end{figure}

\section{Preliminary}

\paragraph{Scaling Laws} Prior research \cite{kaplan2020scalinglawsneurallanguage,hoffmann2022trainingcomputeoptimallargelanguage} indicates that the performance of models and the factors of computing power $C$, dataset $D$, and model parameter $N$ adhere to a power-law relationship. By perpetually augmenting the training dataset and enlarging the model parameter, one can achieve continual enhancements in model performance. The study in \cite{hoffmann2022trainingcomputeoptimallargelanguage} elucidates the boundless capabilities of large-scale models and offers strategic advice on the optimal training of such models under limiting conditions. For instance:
\begin{itemize}
\item $L(N)=(\frac{N_c}{N})^{\alpha _N}$ -- Given a fixed model parameter, if infinite computing power and data are provided, the model can eventually converge to a loss value, which reflects the upper limit of the model's capability under this parameter quantity. $N_c$ and $\alpha _N$ are constant terms that can be obtained through fitting by conducting several experiments on small-scale model sizes. The values of these constants typically depend on vocabulary size and tokenization.
\item $L(D)=(\frac{D_c}{D})^{\alpha _D}$ -- Given a fixed quantity of data, without restrictions on computational power and the number of parameters, the loss function converges to an optimal value, representing the maximum extractable information content from the given data. $L(D)$ denotes the theoretical lower bound of the achievable test loss. In scenarios with limited data and unbounded computational power, the risk of overfitting increases significantly. Consequently, the implementation of an early stopping protocol becomes critical for achieving the global minimum of the test loss function.
\item $L(C)=(\frac{C_c}{C})^{\alpha _C}~(naive)$ -- Assuming a fixed computing power, and without restricting the number of model parameters and data volume, the near-optimal loss attainable is depicted by the equation. The constants $C_c$ and $\alpha_C$ are empirically derived through regression analysis of multiple small-scale model experiments. It is crucial to note that this represents a suboptimal loss, rather than the global minimum. Prior research has established that attaining the optimal loss requires the batch size to satisfy the condition $B << B_{crit}$, where $B_{crit}$ represents a critical threshold value.
\end{itemize}

\paragraph{local SGD}

Stochastic Gradient Descent (SGD) and Adam \cite{kingma2017adammethodstochasticoptimization} are prevalently employed parameter optimization techniques in the domain of deep learning. In large-scale distributed training on clusters, Distributed Data Parallel (DDP) \cite{li2020pytorchdistributedexperiencesaccelerating} and Zero \cite{rajbhandari2020zeromemoryoptimizationstraining}  are prevalent strategies. However, these methods are significantly constrained by network bandwidth limitations. To mitigate the impacts of communication bandwidth constraints, one viable solution is to reduce the volume of communication per iteration through approaches such as quantization algorithms \cite{tang20211bitadamcommunicationefficient}. Another solution is to maintain the volume of communication per iteration while decreasing the frequency of communication. local SGD is an optimization algorithm that effectively reduces the frequency of communication. It operates local SGD independently on various clusters and only occasionally conducts global synchronization.

\cite{douillard2023dilocodistributedlowcommunicationtraining} propose a distributed optimization algorithm that facilitates the training of LLMs on loosely connected islands of devices. This method bifurcates the model training process into two distinct stages. Initially, each cluster updates its model parameters using an inner optimizer, similar to traditional LLM training methods. Subsequently, the discrepancies between their updated model parameters and their initial parameters, then serve as gradients for the outer optimizer. This methodology ensures coherence among the clusters and minimizes the requirements for communication bandwidth.

\paragraph{Notation} We use the following notations, most of which are adapted from \cite{kaplan2020scalinglawsneurallanguage}
\begin{itemize}
\item $L$ -- the cross entropy loss in nats averaged over the tokens in a context
\item $N$ -- the number of model parameters, \emph{excluding all vocabulary and positional embeddings}  
\item $B$ -- the global batch size(tokens)
\item $S$ -- the number of training steps
\item $D$ -- the dataset size in tokens, which $D=BS$
\item $s$ -- local update steps
\item $m$ -- the number of cluster
\item $n$ -- the number of GPU per cluster
\item $W$ -- bandwidth across multi clusters
\item $C_{d}$ -- Floating-point Operations Per Second(FLOPS) per device in practice
\end{itemize}

\begin{figure} 
\noindent \centering{} 
\includegraphics[width=0.32\textwidth]{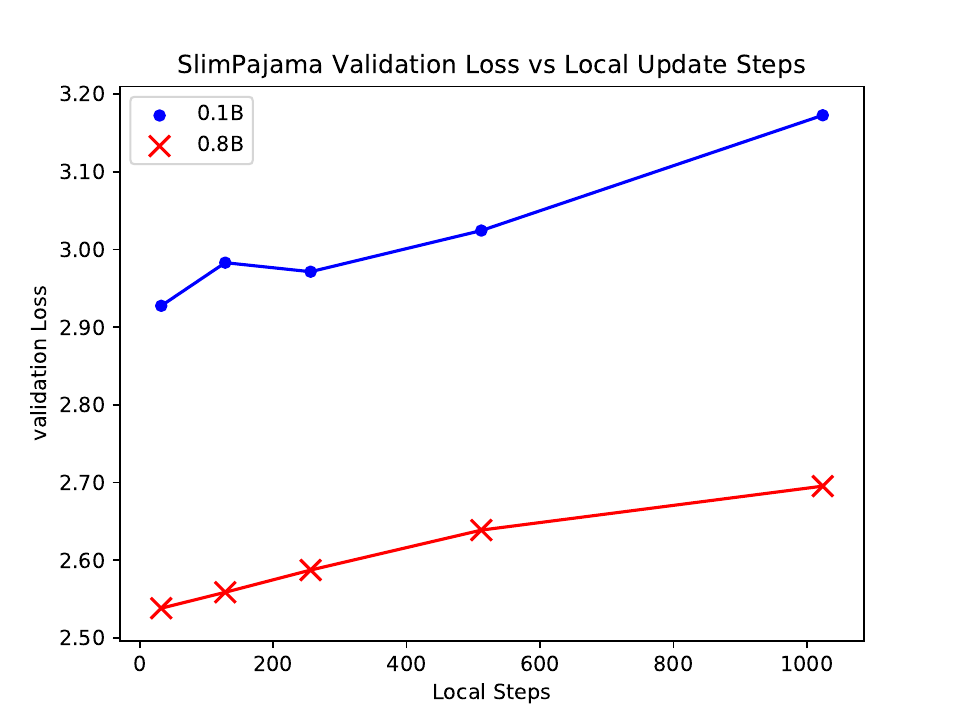}
\includegraphics[width=0.32\textwidth]{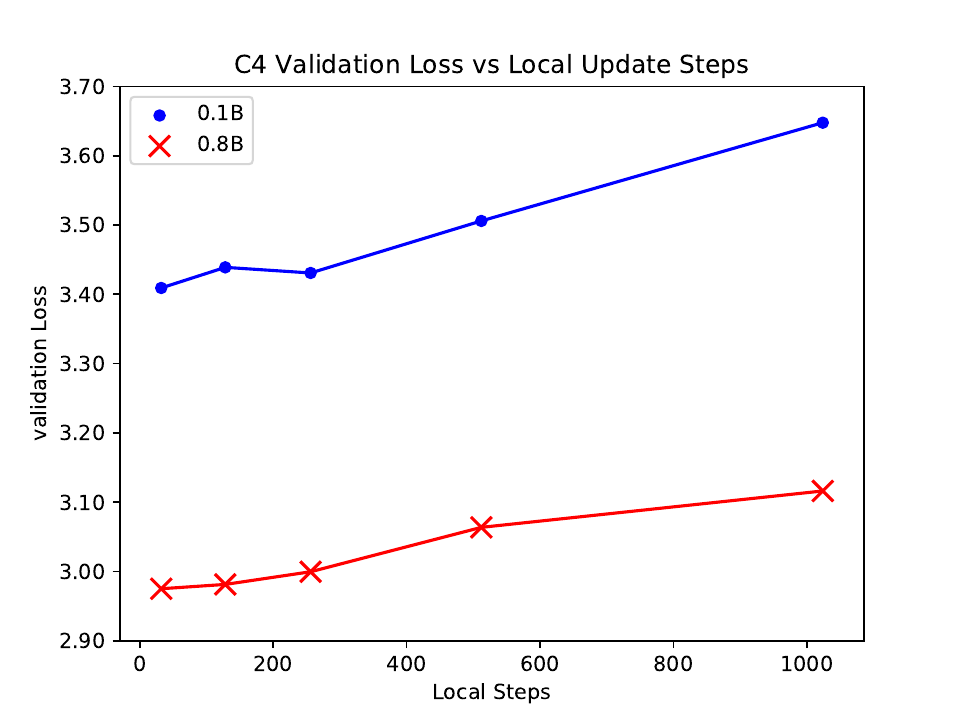}
\includegraphics[width=0.32\textwidth]{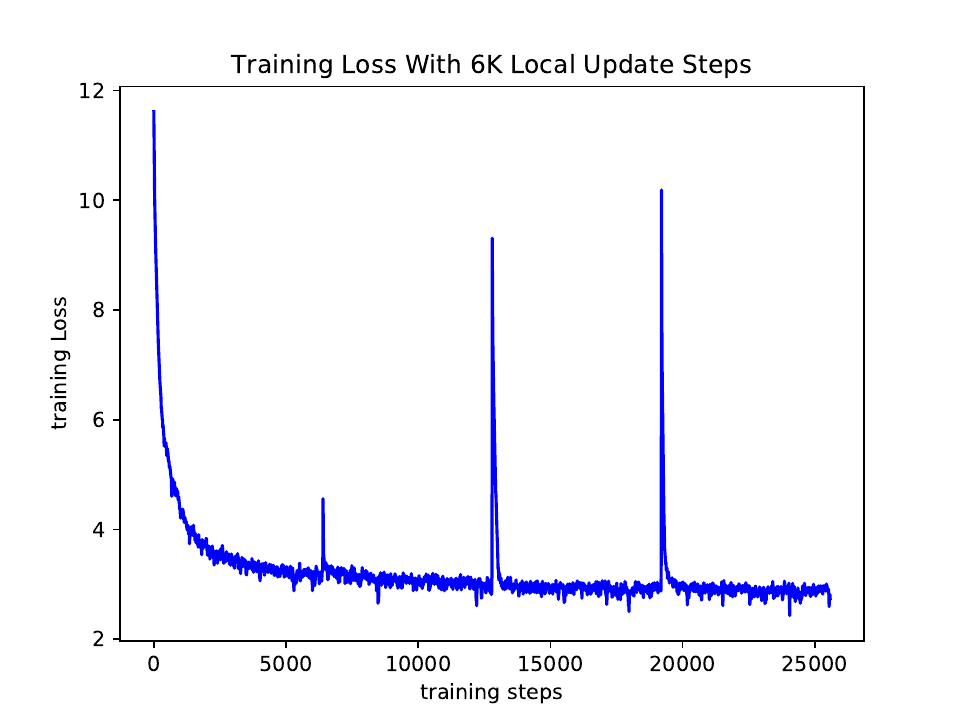}
 \caption[]{Performance of different-sized models on the C4 and SlimPajama datasets as local update steps increase.}
 \label{fig:loss_local_step}
\end{figure}

\section{Scaling Laws for local SGD}
\subsection{Datasets}
\paragraph{Training Data}
In this study, we employed the open-source dataset SlimPajama \cite{cerebras2023slimpajama}, which was developed by CerebrasAI. This dataset possesses high quality and commercial viability, and it has been pivotal in training numerous open-source pre-trained language models \cite{ren2024samba,zhang2024tinyllama,zhuang2024chuxin16btechnicalreport}. SlimPajama is derived from the Red Pajama dataset, initially released by Together AI. Through stringent filtering processes aimed at eliminating duplicate and low-quality data, the dataset was refined from an initial 1.21 trillion tokens to a more focused $627$ billion token.In our research, we utilized the tokenizer from DeepSeek-LLM \cite{deepseekai2024deepseekllmscalingopensource} to preprocess the SlimPajama dataset.
\paragraph{Validation Data} 
To ensure an impartial evaluation of different models and training methodologies, we employed two separate datasets for calculating the generalization error: the validation set of SlimPajama and the C4 dataset \cite{JMLR:v21:20-074}.

The C4 dataset, extracted from the publicly accessible Common Crawl web archive, comprises approximately $750GB$ of English text. Both datasets were subjected to identical preprocessing procedures, consistent with those applied to our training data, to ensure uniformity within our evaluation framework.

Following preprocessing and sampling, the SlimPajama validation set produced an estimated 100 million tokens. Similarly, the C4 dataset, subject to identical preprocessing and sampling procedures, also yielded approximately 100 million tokens. This considerable quantity of evaluation data substantially augments the robustness and reliability of our performance metrics.

\subsection{Training Procedures}

In order to rigorously validate the scaling law of local SGD and draw comparisons with DDP training methodologies, we developed eight distinct model sizes to evaluate the generalization error. Each model was trained to achieve adequate convergence. To corroborate the predictive accuracy of our fitted scaling curves for larger model configurations, we additionally trained a model with 3 billion parameters. To facilitate a fair comparison while ensuring optimal generalization error across all model sizes, we meticulously examined several key hyperparameters. The subsequent section expounds on our considerations and configurations for these critical training hyperparameters.

\paragraph{Batch Size}
To ensure a fair comparison between DDP and local SGD training methods, we maintained an equivalent global batch size, utilizing $4M$ tokens as the standard. In our local SGD experiments, we treated each node (comprising $8$ GPUs) as a local node, ensuring that the number of training nodes in DDP matches that in local SGD. This configuration guarantees consistent training data volume per step between local SGD and DDP, enabling a more equitable comparison. We consistently used $8$ nodes across all experiments. 

\paragraph{Optimizer and Learning Rate}
Following previous LLM training experience, we used the AdamW optimizer for both DDP and local SGD, with $\beta _{1}$ set to $0.9$ and $\beta _{2}$ to $0.95$. We use the cosine decay strategy \cite{loshchilov2017sgdrstochasticgradientdescent} and decay the learning rate to 10\%. For the AdamW optimizer, we conducted a hyperparameter search within a defined range for the learning rate to achieve optimal performance. Following \cite{douillard2023dilocodistributedlowcommunicationtraining} method, we utilized Nesterov momentum as the outer optimizer for local SGD to periodically synchronize models. Based on our observations, the outer optimizer is not sensitive to the learning rate, so we uniformly set the learning rate to 0.7 and momentum to 0.9 without decaying the learning rate.

\paragraph{Local Update Steps}
Local update steps denote the interval at which local SGD synchronizes models. \cref{fig:loss_local_step} suggests that smaller local update steps result in less performance degradation. However, in scenarios with limited cross-node communication bandwidth, very small local update steps may significantly slow down training, which is impractical for real-world applications. Consequently, to strike a balance between computational efficiency and algorithmic effectiveness across various model scales, we uniformly employed 32 local update steps to observe changes in the scaling law of local SGD. A comprehensive examination of how different local update step configurations influence scalability in different environments will be presented in Section \ref{sec: discussion}. In addition, we have experimented with extra-long local update steps, and the results are shown in \cref{fig:loss_local_step}. We found that a significant loss spike was observed after each outer optimizer under 6k local update steps, and then it naturally faded with the inner optimizer, and no significant loss was observed in the model performance after spike regression.

\subsection{Predicting $L(N)$(Non-Embedding)}

In \cref{fig:main_loss} we present the performance of eight models with parameter counts ranging from $5$ million to $800$ million, trained to achieve adequate convergence and validated using the C4 and SlimPajama datasets. Our experiments demonstrate that Distributed Data Parallel (DDP) and local SGD exhibit comparable scaling laws for non-embedding parameter count $N$, which can be characterized by the terms outlined in the following equation:
\begin{equation}
\label{LN_DDP}
    L_{DDP}(N) \approx (\frac{N_c}{N})^{\alpha _N};~~\alpha _N \sim 0.069,~~N_c \sim 6.06 \times 10^{14}
\end{equation}

\begin{equation}
\label{LN_LOCAL}
    L_{localSGD}(N) \approx (\frac{N_c}{N})^{\alpha _N};~~\alpha _N \sim 0.072,~~N_c \sim 3.15 \times 10^{14}
\end{equation}
Besides, we use the test results on the C4 dataset to test its generalization performance. The results indicate that the Out-of-Distribution (OOD) test outcomes exhibit greater consistency compared to the in-distribution test results.

To ensure the accuracy of the scaling law, we further validated it on a 3-billion-parameter model. As shown in \cref{fig:main_loss}, the scaling law accurately predicts the performance of larger models. This indicates that the scaling law has been validated for larger models and computational power, providing strong support for its extension to even larger-scale practical scenarios.

\subsection{Predicting $L(K, N)$(Non-Embedding)} \label{sec: discussion}
\subsubsection{Scaling Efficiency K}

As illustrated in \cref{fig:local_llm}, we consider a distributed system comprising four clusters, each containing $n$ GPUs interconnected via high-bandwidth networks. We designate each independent cluster as a high bandwidth (HB). Within each cluster, we implement 3D or 4D parallelism strategies \cite{narayanan2021efficientlargescalelanguagemodel} to maximize Hardware FLOP Utilization (HFU) and ensure efficient parallel computing. For diagrammatic clarity, we assume that two GPUs can accommodate a complete model, with horizontal scaling achieved through Distributed Data Parallel (DDP) or Zero Optimizer techniques.

Given the Ethernet-based inter-cluster data transmission among the four clusters, we optimize communication efficiency by limiting the synchronization process to a single, complete set of model parameters. This approach minimizes redundant data transfer across the lower-bandwidth Ethernet connections. Following the inter-cluster synchronization, the updated parameters are efficiently disseminated within each cluster via the HB. This two-stage synchronization strategy-inter-cluster via Ethernet followed by intra-cluster via HB networks ensures optimal utilization of available bandwidth resources while maintaining model consistency across the distributed system.

By considering the computational and communicational boundary conditions, we can estimate the minimum communication requirements for cross-regional clusters in realistic scenarios.

The computational cost for a single model parameter update is $6NB$ per step. The intra-cluster computation time per step is given by:
\begin{equation}
\frac{6N\frac{B}{m}}{nC_d}
\end{equation}
Assuming a single GPU can host the entire model, only an $Allreduce$ operation with a communication domain size of $m$ (number of clusters) is necessary. The inter-cluster communication time is expressed as:
\begin{equation}
\frac{2N*2(m-1)}{mW}
\end{equation}
Where $K$ denotes the multi-cluster scaling efficiency, we can formulate the following equation to represent the relationship between these parameters:

\begin{equation}
\begin{aligned}
\label{K}
    K & \approx \frac{s\frac{6N\frac{B}{m}}{nC_{d}}}{\frac{4N(m-1)}{mW}+s\frac{6N\frac{B}{m}}{nC_{d}}}
    \\ & = \frac{1}{1+\frac{2(m-1)nC_{d}}{3BsW}}  ~~ where~ s \leq \frac{D}{B}
\end{aligned}
\end{equation}

\begin{figure} 
\centering
\subfigure[]{\label{Fig.sub.1}
\includegraphics[width=0.48\textwidth]{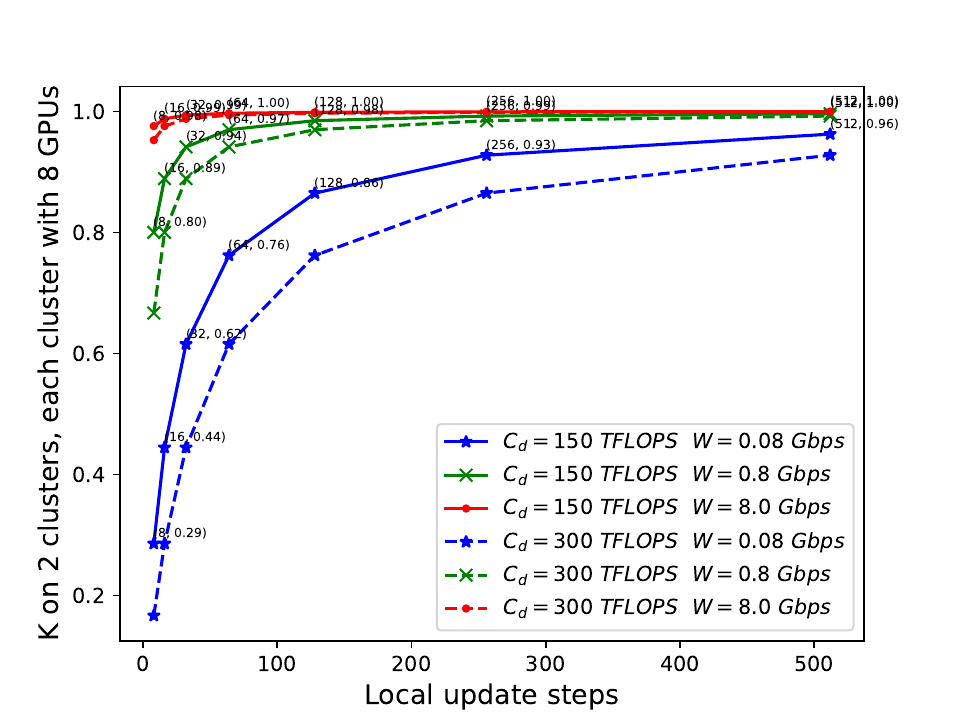}}
\subfigure[]{\label{Fig.sub.2}
\includegraphics[width=0.48\textwidth]{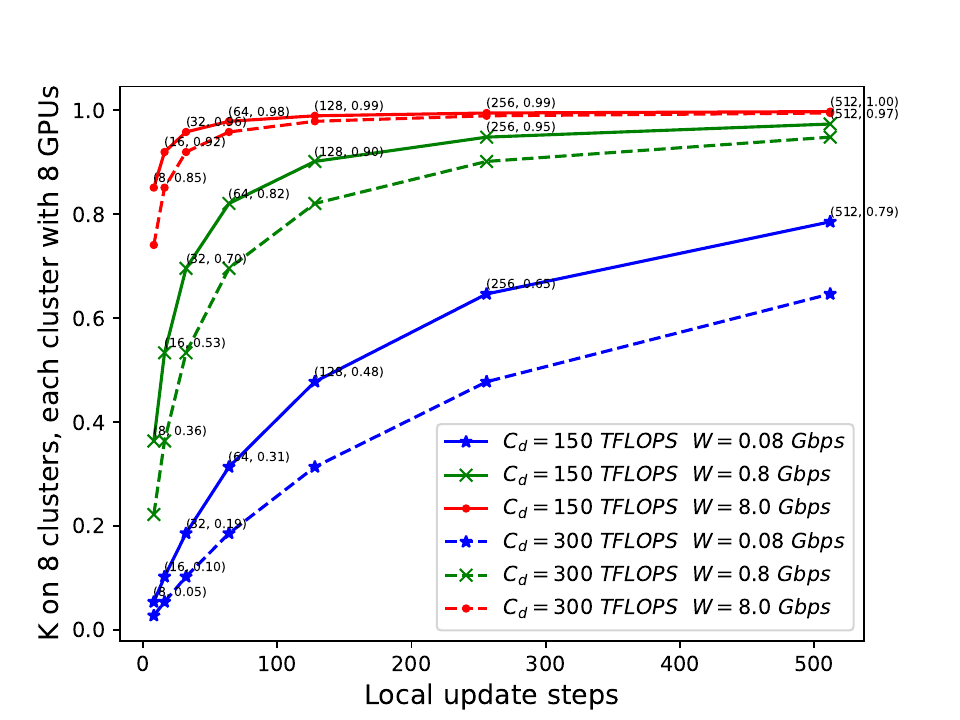}}
\subfigure[]{\label{Fig.sub.3}
\includegraphics[width=0.48\textwidth]{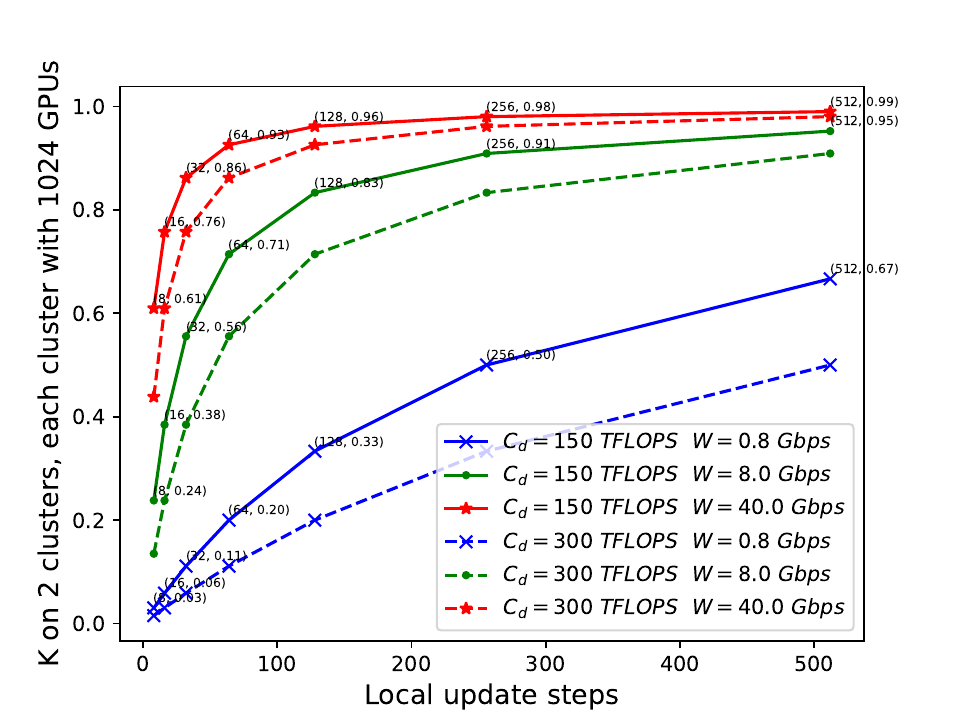}}
\subfigure[]{\label{Fig.sub.4}
\includegraphics[width=0.48\textwidth]{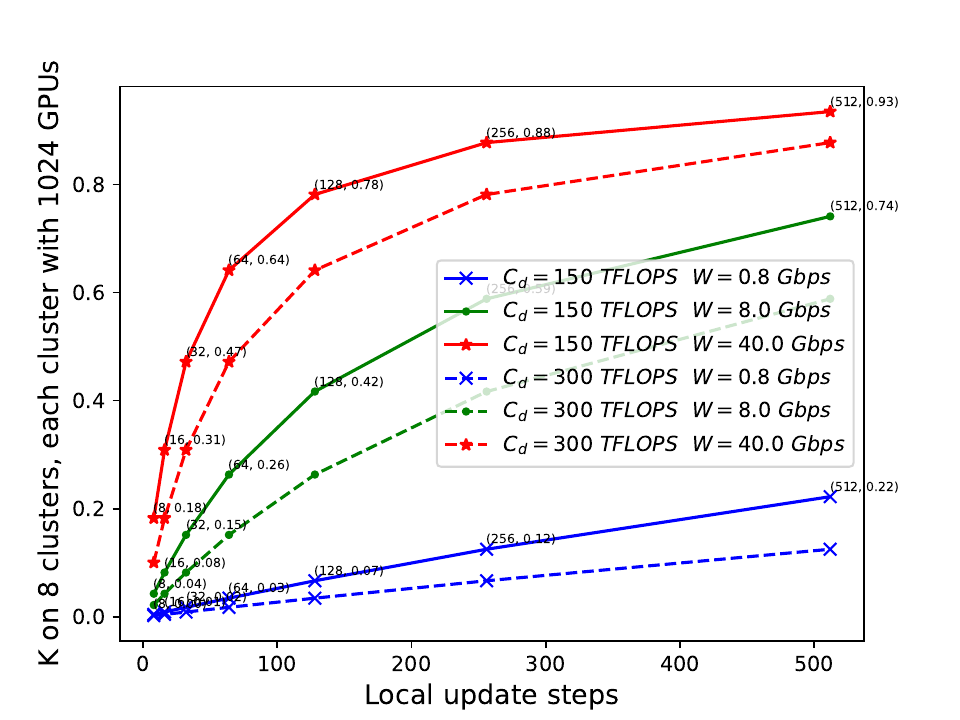}}
\subfigure[]{\label{Fig.sub.5}
\includegraphics[width=0.48\textwidth]{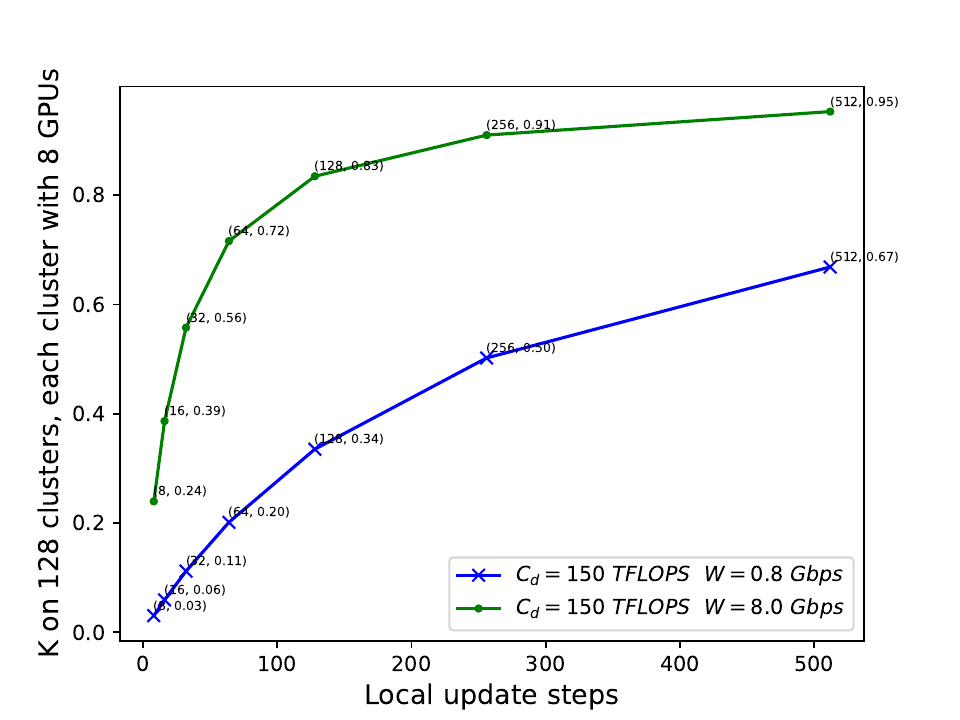}}
\subfigure[]{\label{Fig.sub.6}
\includegraphics[width=0.48\textwidth]{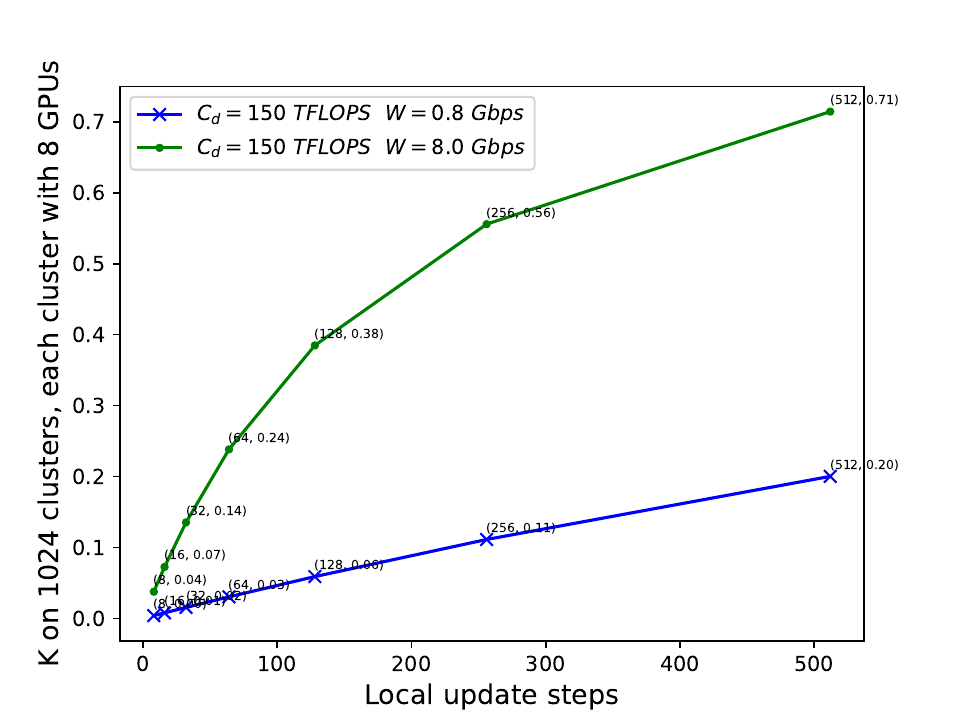}}
\caption[K\_local steps]{$K$ curve with local update steps. $C_d$ denotes FLOPS per device, $W$ denotes Bandwidth}
\label{fig:K_localstep}
\end{figure}

This formulation allows for a more comprehensive analysis of the scaling behavior in multi-cluster, distributed LLM training scenarios.

\paragraph{Scenario 1 Small-Scale Edge Computing}We consider a scenario with $n=8$ and $B=4$ million tokens. We examine how the parameter $K$ varies with computational power $C_d \in \{150, 300\}$ TFLOPS, number of clusters $m \in \{2, 8\}$, and network bandwidth $W \in \{0.08, 0.8, 8.0\}$ Gbps, with results depicted in \cref{Fig.sub.1,Fig.sub.2}. In the context of distributed training systems, our analysis reveals distinct performance patterns based on the $m$ (number of clusters) and available communication bandwidth $W$.

For $m = 2$ (two-cluster configuration):
\begin{itemize}
\item Communication bandwidth $\geq 0.8$ Gbps suffices to maintain high scaling efficiency.
\item When bandwidth is reduced to 0.08 Gbps, a local update step of 64 is recommended to ensure satisfactory parallel efficiency.
\end{itemize}

For $m = 8$ (eight-cluster configuration):
\begin{itemize}
\item With bandwidth $\geq 0.8$ Gbps, local update steps $\geq 32$ can still maintain relatively ideal scaling efficiency.
\item Further reduction in communication bandwidth results in a significant overall decline in scaling efficiency.
\end{itemize}

\paragraph{Scenario 2 Multi-Cluster Training} This scenario considers the scalability across multiple clusters. We assume $n = 1024$ and $B = 4$ million tokens. We explore the relationship between $K$ and $C_d \in \{150, 300\}$ TFLOPS, $m \in \{2, 8\}$, and $W \in \{0.8, 8.0, 40.0\}$ Gbps, with results illustrated in \cref{Fig.sub.3,Fig.sub.4}. 

In contrast to Scenario 1, we observe a significant increase in cross-cluster communication bandwidth requirements as the computational power within individual clusters grows. This relationship highlights the scalability challenges in distributed training systems.

Two-cluster Configuration (1024 GPUs per cluster):
\begin{itemize}
\item Minimum bandwidth requirement: $\geq 8$ Gbps
\item Recommended local update steps: $64$
\end{itemize}

Eight-cluster Configuration (1024 GPUs per cluster):
\begin{itemize}
\item Minimum bandwidth requirement: 40 Gbps(essentially mandatory)
\item Recommended local update steps: 64 or 128
\end{itemize}

Scalability Limitations at the $10,000$ GPU Scale, this analysis reveals substantial constraints on scaling efficiency at the $10,000$ GPU scale. These limitations stem from two primary factors:
\begin{itemize}
\item Communication-Computation Imbalance: In the context of training Large Language Models (LLMs) with approximately $100$ billion parameters, there exists a significant disparity between inter-cluster communication time and intra-cluster computation time. Specifically, the duration required for a single round of communication between clusters is on the order of hours, whereas the time needed for a single computational iteration within a cluster is measured in minutes. This substantial difference in time scales highlights a critical challenge in distributed LLM training across multiple clusters, where the communication overhead can become a dominant factor in the overall training process.

\item Algorithm Constraints: The computational paradigm inherent to the local SGD algorithm precludes the possibility of overlapping computation and communication processes. This inherent limitation results in a significant inefficiency, as computational resources are forced into idle states during inter-cluster communication phases. Consequently, this lack of parallelism between computation and communication leads to suboptimal resource utilization. 
\end{itemize}

\paragraph{Scenario 3 Large-Scale Edge Computing} We assume $n = 8$ and $B = 4$ million tokens. We examine the relationship between $K$ and $C_d \in \{150\}$ TFLOPS, $m \in \{1024\}$, and $W \in \{0.08, 0.8\}$ Gbps, with results presented in \cref{Fig.sub.6}. As $m$ increases, the dependence on cross-cluster bandwidth $W$ similarly intensifies. This trend closely aligns with Scenario 2. However, it is crucial to note that in edge computing scenarios, ensuring high bandwidth for each small-scale cluster is challenging, presenting significant practical implementation difficulties.

\subsubsection{$L(K, N)$(Non-Embedding)}
\cref{fig:loss_local_step} presenting the performance of different-sized models on the C4 and SlimPajama datasets as local update steps increase. We find that there is a linear relationship between $L$ and $N,s$, and when combined with \cref{LN_DDP,LN_LOCAL}, we get the following formula:

\begin{equation}
\begin{aligned}
\label{L_s_N}
    L(s, N) & \approx \alpha _s s + L(N) ~~~~~where~s < 1024
\end{aligned}
\end{equation}

combine \cref{LN_LOCAL,K,L_s_N}:

\begin{equation}
\begin{aligned}
\label{L_K_N}
    L(K, N) & \approx \alpha _s \frac{2KnC_d(m-1)}{3(1-K)WB} + (\frac{N_c}{N})^{\alpha _N}
    \\ & = \lambda \frac{K}{1-K} + (\frac{N_c}{N})^{\alpha _N} ~~~~~ where ~~ \lambda=\frac{2\alpha _s nC_d(m-1)}{3WB}
\end{aligned}
\end{equation}

Analysis of $\lambda$ in \cref{L_K_N}. Under the following conditions:
\begin{itemize}
\item utilizing $\alpha_s$ derived from \cref{fig:loss_local_step}
\item $n,m = 8$ (8 GPUs per cluster, 8 clusters)
\item $C_d = 150$ TFLOPS (per device in practice)
\item $W = 0.8$ Gbps (inter-cluster bandwidth)
\item $B = 4$ million tokens (batch size)
\end{itemize}
Our calculations yield $\lambda \approx 2*10^{-3}$, a notably small constant. This result provides additional validation from an alternative perspective for the approximation between $L_{DDP}(N)$ and $L_{localSGD}(N)$. The negligible magnitude of $\lambda$ supports the hypothesis that the loss functions for Distributed Data Parallel (DDP) and local SGD converge under these specific conditions. This convergence implies that, in this scenario, the performance difference between these two distributed training approaches is minimal.

\section{Limitation and Future Work}

\paragraph{Limited Computational Resources} It is important to note that our experimental findings were obtained within the constraints of limited computational resources. Furthermore, the experimental results for Scenarios 2 and 3, which involved nearly $10,000$ GPUs, still lack empirical validation in practical settings.

\paragraph{Optimizer} Our validation of the scaling law employed AdamW as the inner optimizer and Nesterov Momentum as the outer optimizer, which yielded optimal results. However, model convergence can be achieved with alternative optimizers. Further research into the impact of diverse optimizers on the scaling law presents an avenue for future investigation.

\paragraph{Model Architecture} While our experiments primarily focused on transformer-based language models, numerous competitive model architectures exist. Exploring the applicability of the scaling law to these alternative structures represents a promising research direction.

\paragraph{Quantization and Sparsification} In \ref{sec: discussion}, We did not explicitly address quantization and sparsification techniques \cite{alistarh2018convergencesparsifiedgradientmethods,shi2019distributedsynchronoussgdalgorithm} for improving communication latency. We posit that these methods, while potentially enhancing communication efficiency, do not fundamentally alter the validity of the scaling law. Moreover, we believe that these techniques, despite their ability to significantly reduce communication latency, do not address the core efficiency bottlenecks inherent in large-scale local SGD training.

\paragraph{Critical Batch Size} In \ref{sec: discussion}, our scaling law analysis assumed a global batch size of 4 million tokens. Based on our limited experience, excessively large global batch sizes can reduce computational efficiency in LLM training. We hypothesize that $B_{crit}$ remains relatively constant in the context of local SGD. However, a significant increase in $B_{crit}$ for local SGD could substantially reduce the proportion of communication latency. This possibility warrants further exploration in future studies.

\paragraph{Heterogeneous Cluster Configurations} Our current analysis does not account for scenarios where different clusters possess varying numbers of GPUs, leading to disparities in data processing volumes across clusters. The potential impact of this heterogeneity on the outer optimizer's update strategy remains an open question, necessitating further investigation.

\section{Conclusion}
This paper presents a comprehensive analysis of scaling laws for local SGD in Large Language Model training. Our research addresses the growing challenges in the computational power that arise as LLM parameters continue to increase. By investigating alternative optimization algorithms, particularly local SGD, we offer insights into more efficient and scalable approaches for LLM training.

Furthermore, our research opens new avenues for training LLMs on loosely connected clusters or edge devices, potentially democratizing access to large-scale AI model training. The insights gained from our analysis of multi-cluster and edge computing scenarios offer valuable guidance for researchers and practitioners aiming to push the boundaries of LLM scale and efficiency.

In conclusion, this work not only advances our understanding of scaling laws in distributed LLM training but also provides practical insights for the design and implementation of future large-scale AI systems. As LLMs continue to grow in size and importance, the methodologies and insights presented in this paper will play a crucial role in shaping the future landscape of AI research and application.
\bibliographystyle{halpha}
\bibliography{main}
\end{document}